\documentclass{bmvc2k}
\title{On Automatic Data Augmentation for 3D Point Cloud Classification}

\addauthor{Wanyue Zhang$^*$}{wzhang@mpi-inf.mpg.de}{1}
\addauthor{Xun Xu$^\dagger$}{xux@i2r.a-star.edu.sg}{1}
\addauthor{Fayao Liu}{liu_fayao@i2r.a-star.edu.sg}{1}
\addauthor{Le Zhang$^\ddagger$}{zhangleuestc@gmail.com}{1}
\addauthor{Chuan-Sheng Foo}{foo_chuan_sheng@i2r.a-star.edu.sg}{1}

\addinstitution{
 Institute for Infocomm Research\\
 A*STAR, Singapore
}

\runninghead{Zhang Et Al}{Automatic data augmentation for 3D point cloud}

\usepackage{amsmath}
\usepackage{tabularx}
\usepackage{array}
\usepackage[ruled,vlined]{algorithm2e}

\usepackage{booktabs, multirow} 
\usepackage{rotating} 
\usepackage{capt-of}
\usepackage{amssymb}

\newcommand{\vect}[1]{\mathbf{#1}}
\newcommand{\matr}[1]{\mathbf{#1}}

\newcommand{\set}[1]{\mathcal{#1}}

\begin{document}

\maketitle

\begin{abstract}

Data augmentation is an important technique to reduce overfitting and improve learning performance, but existing works on data augmentation for 3D point cloud data are based on heuristics. In this work, we instead propose to automatically learn a data augmentation strategy using bilevel optimization. An augmentor is designed in a similar fashion to a conditional generator and is optimized by minimizing a base model's loss on a validation set when the augmented input is used for training the model. This formulation provides a more principled way to learn data augmentation on 3D point clouds. We evaluate our approach on standard point cloud classification tasks and a more challenging setting with pose misalignment between training and validation/test sets. The proposed strategy achieves competitive performance on both tasks and we provide further insight into the augmentor's ability to learn the validation set distribution.
\end{abstract}


\section{Introduction}

\label{sec:intro}
\let\thefootnote\relax\footnote{$\dagger$ Correspondence to Xun Xu. $^*$ Wanyue Zhang is now with the Max Planck Institute for Informatics. $\ddagger$ Le Zhang is now with the University of Electronic Science and Technology of China.}

Understanding 3D point clouds is crucial to a wide range of applications including autonomous driving, robotics and human-computer interaction. Recent progress in this area is attributed to training deep neural networks on labeled data \cite{qi2017pointnet,qi2017pointnet++}. In order to increase the diversity of training data to aid generalization, data augmentation has been used in the training of neural networks for 3D point clouds to much success \cite{chen2020pointmixup,li2020pointaugment,zhang2021pointcutmix}. Current data augmentation methods for point clouds can be broadly classified into heuristic methods and learning-based methods. In the former category, mixup \cite{zhang2017mixup, lee2021regularization, chen2020pointmixup} between multiple shapes is often used to synthesize novel training examples. Despite being able to provide consistent performance improvements, potentially due to stronger regularization with label smoothing, the data sample after mixup often does not resemble a realistic object. For instance, when applied to a plane and a car, mixup does not yield a semantically meaningful object; the reason behind the success of mixup remains unclear. Learning-based augmentation methods, on the other hand, automatically learn an augmentation strategy instead of hand-engineering one. PointAugment \cite{li2020pointaugment} trains a neural network as the augmentor to predict the rotation and additive jittering parameters for each sample. A heuristic learning objective encouraging augmented samples to be different from the original is used to update the parameters of the augmentor. While PointAugment has showed improved performance over baseline methods, its learning objective does not necessarily guarantee that the augmented sample will benefit the main task. 

By contrast, in the image domain, recent automatic data augmentation methods adopt a bilevel optimization framework
\cite{cubuk2018autoaugment,li2020differentiable,NEURIPS2019_6add07cf} where the learnt augmentation strategy minimizes a base model's loss on validation data while the model is trained using this strategy. Solving the bilevel optimization problem then involves iterating between training the base model and updating augmentation policies. 
Inspired by the recent success of bilevel optimization for automatic data augmentation on images \cite{li2020differentiable,NEURIPS2019_6add07cf}, we propose a novel 3D point cloud augmentation method, \textbf{A}utomatic \textbf{d}ata \textbf{a}ugmentation for 3D \textbf{P}oint \textbf{C}loud data (AdaPC). 
We define the augmentation procedure as sampling from a sequence of transformation operations each parameterized by a distribution, then apply the transformations to each training sample. 
To allow for a high degree of flexibility for augmentation, we use a neural network termed as the \emph{augmentor} to model such distributions. The augmentor parameters are then learned using a bilevel optimization framework by minimizing a base classifier's loss on validation data when the model is trained using the augmentor.
For computational efficiency, the hypergradients, i.e the gradients of the augmentor parameters  w.r.t. the validation loss, are approximated by a one-step unrolling algorithm \cite{liu2018darts} and augmentor parameters are updated with gradient descent. 

AdaPC has several advantages over existing methods for data augmentation on 3D point clouds. First, compared to heuristic approaches \cite{chen2020pointmixup, zhang2021pointcutmix}, AdaPC is able to learn the optimal augmentation for a particular dataset as it directly minimizes the validation loss, while heuristic approaches may only work well in certain cases and require extensive manual hyperparameter tuning. Second, compared to PointAugment \cite{li2020pointaugment}, AdaPC optimizes the augmentor in the direction that minimizes validation error which makes it more likely to generalize to test data. More importantly, if the training set distribution does not exactly match that of the validation/test set (e.g. due to pose mismatch), AdaPC can learn to bridge the gap by capturing any misalignment with the augmentor; we demonstrate that AdaPC is indeed superior in an experimental setup with non-fixed pose. Insight into the effectiveness of the augmentor is also provided by analyzing the distributions learnt by the augmentor.



We summarize the contributions of this work below:
\begin{itemize}
    \item We propose an automatic data augmentation method for 3D point cloud classification by training the augmentor and classifier jointly through bilevel optimization. Compared to existing augmentation methods on 3D point clouds, we provide a more principled way for learning data augmentation strategies.
    \item We achieve superior results on two standard 3D point cloud classification benchmarks compared to other data augmentation methods.
    \item We demonstrate that our augmentor is able to learn meaningful augmentation policies when there is a pose mismatch between training and validation/test sets and provide insight into the distribution of augmentation learned under this scenario.
\end{itemize}

\vspace{-0.3cm}

\section{Related work}

\subsection{Data Augmentation for 3D Point Clouds}

Deep learning methods for 3D point clouds~\cite{qi2017pointnet, qi2017pointnet++, thomas2019kpconv, phan2018dgcnn, atzmon2018point} commonly incorporate small perturbations such as scaling, jittering, point dropout, flipping and rotation to enhance the diversity of the training set. Recently, more sophisticated data augmentation strategies have been studied. Inspired by the success of mixing up images as augmentation~\cite{zhang2017mixup, verma2019manifold},  PointMixUp~\cite{chen2020pointmixup} proposed to interpolate two point clouds by finding a shortest path between the two shapes on the manifold. PointCutMix~\cite{zhang2021pointcutmix} further extended CutMix~\cite{yun2019cutmix} by first finding correspondences between 2 point clouds and then swapping selected regions. Despite demonstrating improvements on standard point cloud datasets, both works are based on heuristics and do not explain the success of the proposed methods. Orthogonal to heuristic augmentation strategies, PointAugment~\cite{li2020pointaugment} formulates augmentation as learning an augmentor which is reminiscent of PointNet. However, it employs a hand-engineered adversarial loss function to train the augmentor which does not necessarily guarantee that the learned augmentation is beneficial and often leads to unstable results. For 3D object detection, PA-AUG~\cite{choi2020part} divides object bounding boxes into fixed partitions and stochastically applies 5 basic transformations such as noise, points dropout, sampling, cutmix~\cite{yun2019cutmix}, cutmixup~\cite{yoo2020rethinking}. Progressive population based augmentation (PPBA)~\cite{cheng2020improving} uses evolutionary search to narrow down the search space of optimal augmentation policies.
In this work, we propose to formulate learning data augmentation in a more principled way. The base model's loss on a validation set is exploited to guide the learning of augmentation strategies that results in improved performance and better generalization to more challenging scenarios.
\vspace{-0.3cm}


\subsection{Automatic Data Augmentation}
Hand-crafted augmentation rules can be ineffective in selecting the optimal combination of augmentation policies as well as their magnitudes. SmartAugment~\cite{lemley2017smart} tackles the problem by training an augmentor to combine samples from the same class alongside the target network. There is also a line of GANs-based approaches~\cite{mun2017generative, perez2017effectiveness, tanaka2019data, antoniou2017data} which generate augmented samples according to the feedback from the main task.  
Another class of methods learns an augmentation strategy based on the validation set.
AutoAugment~\cite{cubuk2018autoaugment} adopts reinforcement learning (RL) to select augmentation policies which improve validation accuracy. 
RandAugment~\cite{cubuk2020randaugment} eliminates the search phase used by previous automatic augmentation methods and conducts a grid search to find the augmentation policies. In light of slow convergence of RL-based approaches, FastAutoAugment~\cite{NEURIPS2019_6add07cf} uses Bayesian optimization to efficiently learn augmentation strategies. DADA~\cite{li2020differentiable} further boosted efficiency by introducing a gradient-based policy searching paradigm and employed bilevel optimization to update augmentation policies. 
DABO~\cite{mounsaveng2021learning} proposed a differentiable augmentor which is capable of generating affine and color transformations. MetaMixUp~\cite{mai2021metamixup} focused on finding a dynamic and adaptative interpolation ratio in an online fashion. 
In this work, we take a similar approach to optimize the augmentor based on the validation performance using bilevel optimization. For computational efficiency, the hypergradients are approximated using the algorithm described in~\cite{liu2018darts}.

\begin{figure}[!htb]
\begin{center}
    \includegraphics[width=0.85\linewidth]{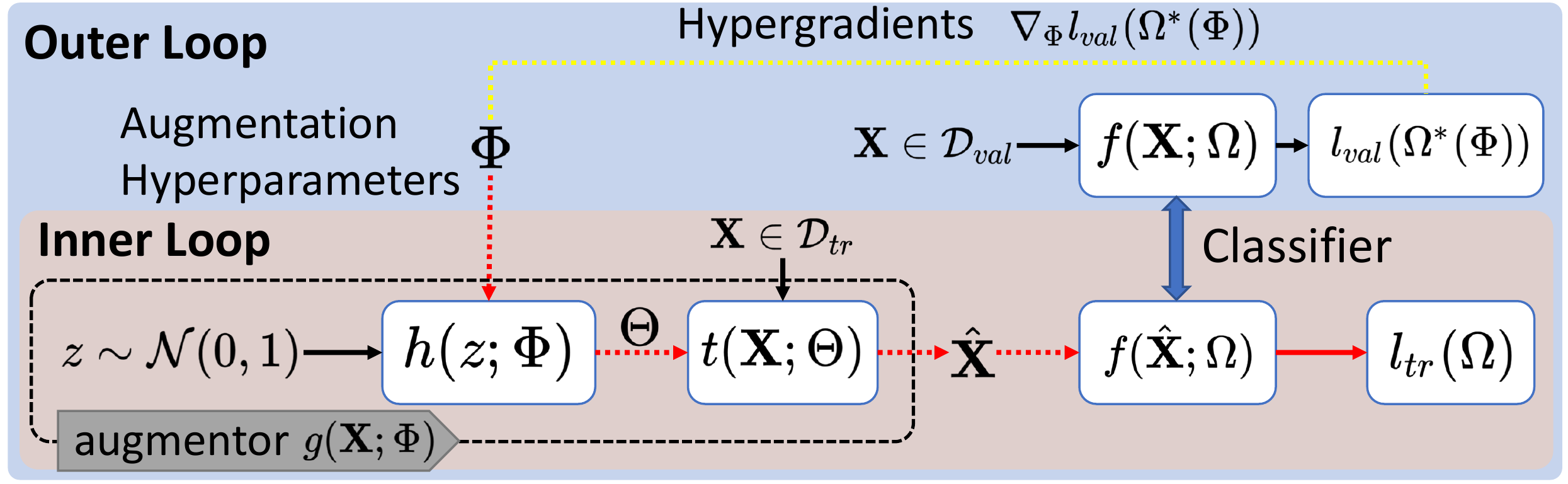}
\end{center}
\caption{Bilevel optimization framework for automatic data augmentation. The solid red arrow indicates a forward pass in the inner loop. The red dashed arrow indicates the update path for outer loop optimization. We use the yellow dashed line to highlight that the hypergradients are approximated instead of being computed by automatic differentiation.
}
\label{fig:main}
\end{figure}

\section{Methodology}

We first provide an overview of AdaPC in Fig.~\ref{fig:main}. During training, AdaPC iterates between an inner loop and outer loop optimization. The inner loop optimization involves updating the base classifier network $f(\cdot;\Omega)$ by minimizing the training loss $l_{tr}$. The forward pass for inner loop optimization is indicated by the solid red arrow. The outer loop optimization updates the data augmentation parameters (or hyperparameters) $\Phi$. Since the validation loss $l_{val}(\Omega^*(\Phi))$ implicitly depends on $\Phi$, we approximate the hypergradients following~\cite{liu2018darts}. The outer loop update path is indicated by the red dashed arrows. In the following section, we first introduce our bilevel optimization framework in Sect.~\ref{sect:bilevel} followed by the specific designs of the augmentor in Sect.~\ref{sect:augmentor}

\subsection{Learning Data Augmentation with Bilevel Optimization}\label{sect:bilevel}

We introduce the bilevel optimization framework for learning data augmentation. We define the augmentation as a mapping $g:\matr{X}\rightarrow\hat{\matr{X}}$ parameterized by $\Phi$ (the augmentation hyperparameters), which is to be learned. This augmentation often involves randomness with instantiations explained in Sect.~\ref{sect:augmentor}.  Let $\set{D}_{tr}=\{\matr{X}_{tr},y_{tr}\}$ and $\set{D}_{val}=\{\matr{X}_{val},y_{val}\}$ denote training and validation sets, $f:\matr{X}\rightarrow y$ parameterized by $\Omega$ denote the classification network, $l_{val}(\Omega;\set{D}_{val})=\frac{1}{|\set{D}_{val}|}\sum_{\matr{X}_i,y_i\in\set{D}_{val}}CE(f(\matr{X}_i;\Omega),y_i)$ denote the validation loss and $l_{tr}(\Omega,\Phi;\set{D}_{tr})$ 
denote the training loss. Since augmentation is turned off during validation, $l_{val}$ does not explicitly depend on $\Phi$. We now formally write the bilevel optimization objective in Eq.~\ref{eq:bilevel} where the left problem is often called the outer loop and the right problem is the inner loop. 
\vspace{-0.2cm}

\begin{equation}\label{eq:bilevel}
    \min_{\Phi} l_{val}(\Omega^*(\Phi);\set{D}_{val}),\quad s.t.\quad \Omega^*(\Phi)=\arg\min_{\Omega}l_{tr}(\Omega,\Phi;\set{D}_{tr})
\end{equation}

The above objective aims to find the optimal augmentation hyperparameters $\Phi$ such that when the model is trained with $\Phi$ (to obtain model parameters $\Omega^*$) its loss on the validation data is minimized. If we assume that the validation set has the same distribution as the test set, this formulation is more principled than PointAugment~\cite{li2020pointaugment} in guaranteeing better generalization. Similar optimization problems have been formulated for neural architecture search~\cite{liu2018darts} and automatic data augmentation~\cite{li2020differentiable}. Solving the inner loop until convergence (training the model to convergence) followed by updating the outer loop leads to a very inefficient optimization paradigm as the inner loop usually takes many steps to converge. We therefore instead used a more efficient approximation algorithm proposed in \cite{liu2018darts}.

Let $\xi$ denote the learning rate for the inner loop optimizer. One step of gradient descent on the classification network parameters can then be written as $\Omega^{'} = \Omega - \xi \nabla_\Omega l_{tr}(\Phi)$.
By approximating the best response (optimal) model parameters $\Omega^*$ to hyperparameters $\Phi$ with parameters obtained from one step of gradient descent $\Omega^\prime$, we obtain approximate hypergradients using the chain rule as follows:
\begin{equation}
\nabla_{\Phi} \ l_{val}(\Omega^{*}(\Phi)) \approx \nabla_{\Phi} \ l_{val}(\Omega^{'}(\Phi))
= - \xi \nabla^2_{\Phi, \Omega} l_{tr}(\Omega, \Phi)\nabla_{\Omega^{'}}l_{val}(\Omega^{'})
\end{equation}

We further use a central finite difference to approximate the Hessian-vector product. Let $\epsilon$ be a small perturbation, $\Omega^{+} = \Omega + \epsilon \nabla_\Omega{'}l_{val}(\Omega')$ and $\Omega^{-} = \Omega - \epsilon \nabla_\Omega{'}l_{val}(\Omega')$. Then, 

\begin{equation}
- \xi \nabla^2_{\Phi, \Omega} l_{tr}(\Omega, \Phi)\nabla_{\Omega^{'}}l_{val}(\Omega^{'})
\approx - \xi \frac{\nabla_{\Phi} l_{tr}(\Omega^{+}, \Phi) - \nabla_{\Phi}l_{tr}(\Omega^{-}, \Phi)}{2 \epsilon}.
\end{equation}

We follow DARTS~\cite{liu2018darts} to set $\xi$ to the learning rate of the classifier and $\epsilon$ to $\frac{0.01}{\lVert \nabla_\Omega{^{'}}l_{val}(\Omega^{'})\lVert} $.
Since the hypergradients tend to be noisy if the hyperparameters are initialized far from a local optimum, we include an L2 regularization term, $l_{reg} = \lVert \Theta - \hat{\Theta} \lVert^2_2$, to the output of the augmentation network to smooth the hypergradients. Prior knowledge $\hat{\Theta}$ about the hyperparameters can be incorporated using this regularizer, e.g. for scaling, $\hat{\theta}=1$ and for rotation, $\hat{\theta}=0$. 
Overall, the final hypergradients with regularization are given by Eq.~\ref{eq:finalgrad} where $\lambda$ is the weight of the regularization term, and the training procedure is presented in Algo.~\ref{alg:bilevel}.

\vspace{-0.2cm}

\begin{equation}\label{eq:finalgrad}
\nabla_{\Phi} l_{val}(\Omega^{*}(\Phi)) \approx - \xi \nabla^2_{\Phi, \Omega} l_{tr}(\Omega, \Phi)\nabla_{\Omega^{'}}l_{val}(\Omega^{'})
 + 2\lambda (\Theta-\hat{\Theta})
\end{equation}

\begin{algorithm}[H]
\SetAlgoLined
 Initialize classifier weights $\Omega$ and augmentor hyperparameters $\Phi$\;
 \While{not converged}{
      
One-step inner loop update $\Omega' = \Omega - \xi \nabla_\Omega l_{tr}(\Phi)$\;
    One-step outer loop update $\Phi' = \Phi -  \alpha\nabla_{\Phi}l_{val}(\Omega^{'}(\Phi))$ \;
    Update classifier weights $\Omega= \Omega - \xi \nabla_\Omega l_{tr}(\Phi^{'})$\$
  
 }
 \Return{$\Omega, \Phi$}
 \caption{Algorithm for learning model weights $\Omega$ and augmentor weights $\Phi$}
 \label{alg:bilevel}
\end{algorithm}
\vspace{-0.3cm}

\subsection{Augmentation Operations}\label{sect:augmentor}

In this section, we introduce the design of an augmentor which can be easily incorporated into the bilevel optimization framework.
Inspired by recent studies into deep learning on 3D point clouds \cite{li2020pointaugment}, we apply two types of augmentation operations. First, we apply a point-wise jittering  $\matr{J}\in\mathbb{R}^{N\times 4}$ (in homogeneous coordinates) which can be seen as a deformation of the original 3D shape. To limit the learnable parameters for jittering, we sample jittering from a uniform distribution $\mathcal{U}(\theta_J,\theta_J+\upsilon)$ where $\theta_J$ is a learnable parameter and $\upsilon$ determines a fixed range. Then, we further apply a rigid transformation, represented in homogeneous coordinates as $\matr{H}\in\mathbb{R}^{4\times4}$ to the perturbed 3D shape. Instead of directly regressing a transformation matrix like PointAugment, we parameterize $\matr{H}=\matr{T}\cdot\matr{R}\cdot\matr{S}$ where $\matr{R}$ is a rotation matrix parameterized by $\vect{\theta_r}\in\mathbb{R}^{1}$, $\matr{S}$ is a scaling matrix parameterized by $\vect{\theta_s}\in\mathbb{R}^{3}$ and $\matr{T}$ is a translation matrix parameterized by $\vect{\theta_t}\in\mathbb{R}^{3}$. Such a decomposition allows the augmentor to directly regress individual parameters; as a result we can easily apply regularization to individual operations and each operation can be randomly dropped out to create more diverse augmentations. In contrast, it is difficult to incorporate explicit regularization when directly regressing onto $\matr{H}$. 
For brevity, we denote $\Theta=[\theta_J,\vect{\theta_r},\vect{\theta_s},\vect{\theta_t}]\in\mathbb{R}^{8}$ and the transformation operation is described by Eq.~\ref{eq:aug} as follows:
\vspace{-0.3cm}

\begin{equation}\label{eq:aug}
    t(\matr{X};\Theta)=(\matr{X}+\matr{J}(\Theta))\cdot \matr{H}(\Theta)^\top.
\end{equation}

One approach to obtain $\Theta$ is to learn a fixed vector that applies to all training examples. However, this does not allow randomness in the augmentation procedure. Another approach is to model $\Theta$ using a parametric distribution $p(\Theta)$, e.g. Gaussian or uniform distribution, which is easy to parameterize and sample from. For example, if a Gaussian distribution is used, we parameterize it as $\theta_i\sim\mathcal{N}(\mu,\sigma)$ where $\mu$ and $\sigma$ are hyperparameters to optimize and $\Omega=[\mu,\sigma]$. We discuss such augmentor designs in more detail in Section S1 of the supplementary material. Despite being more flexible, it is not trivial to specify the most appropriate distribution without any prior knowledge. To further increase the flexibility of the distribution used to model $\Theta$, we propose to use a neural network denoted as $h(z;\Phi)$, as illustrated in Fig.~\ref{fig:main}. The neural network takes a random noise vector with components sampled from a zero-mean, unit variance Gaussian distribution $z\sim\mathcal{N}(0,1)$ and transforms it into augmentation parameters $\Theta=h(z;\Phi)$. This network is similar to a generator in a GAN~\cite{goodfellow2014generative} which maps a simple distribution to a more complicated one. The generator is trained by optimizing the performance on the validation set.
Overall, the augmentor $g(\matr{X};\Phi)$ can be seen as a conditional generator~\cite{mirza2014conditional} and augmentation is regarded as a conditional generative process.

\vspace{-0.5cm}

\section{Experiments}
\vspace{-0.3cm}
\subsection{Datasets}
We evaluate on \textbf{ModelNet40~\cite{wu20153d}} for shape classification. It consists of 12311 CAD models from 40 categories of which 9843 models are used for training and 2468 for testing. We use a pre-processed dataset which contains uniformly sampled points from the mesh surfaces as input to our network. Following the standard practice, we use 1024 points per shape. 

To test the robustness of our method on objects cropped from real 3D scans, we use the OBJ\_ONLY variant provided by \textbf{ScanObjectNN~\cite{uy2019revisiting}} without the background. OBJ\_ONLY contains objects with noisy surfaces of non-uniform density from 15 categories. 

Our method requires a validation set which is not provided by the above 2 datasets. Hence, we use 90\% of the original training data as the new training data, and the remaining 10\% as the validation data. To ensure a fair comparison with other methods, we randomly re-split the training and validation set every epoch so that that our network has access to the entire training data across epochs.

\vspace{-0.3cm}



\subsection{Implementation details}

The initial learning rate for both classifier $\xi$ and augmentor $\alpha$ are 0.001 with Adam optimizer ($\beta_1 = 0.9$, $\beta_2 = 0.999$). The classifier learning rate is halved every 20 epochs. The classifier and augmentor are updated every minibatch and we train for 300 epochs in total. We use a batch size of 24 for PointNet and 12 for PointNet++ and DGCNN. The regularization weight $\lambda$ is a manually tunable hyperparameter. A proper selection of $\lambda$ is vital to the stability of training. We use $\lambda=0.5$ for ModelNet40 and $\lambda=10.0$ for ScanObjectNN. We use cross-entropy loss for classification. We adopt other training practices from PointAugment for fair comparison, e.g. adding a term to penalize the feature difference between the original and augmented samples, apply dropout to the augmentation types and mix original data and augmented data during training. 
\vspace{-0.3cm}


\subsection{Automatic Augmentation for Classification}\label{sect:ClassExp}
We first report results on standard benchmark datasets for classification on point clouds, ModelNet40 and ScanObjectNN using PointNet as the backbone~\cite{qi2017pointnet} in Tab.~\ref{tab:Classification}. We compare the following approaches in this experiment. First, we evaluate three baseline methods with a fixed predefined augmentor, referred to as Fixed Aug. in Tab.~\ref{tab:Classification}. Vanilla PointNet without any augmentation, PointNet with predefined scaling and translation as augmentation, referred to as Sampling-S,T in Tab.~\ref{tab:Classification} and PointNet with predefined scaling, translation and rotation, referred to as Sampling-S,T,R are respectively evaluated. For predefined scaling (S), we adopt the practice in PointAugment~\cite{li2020pointaugment} by randomly sampling from a uniform distribution $\mathcal{U}(0.67,1.5)$. For predefined translation (T), we uniformly sample from $\mathcal{U}(-0.2,0.2)$. For predefined rotation about the y-axis (Ry), we sample from $\mathcal{N}(0,0.06^2)$ with output clipped at $0.18$. For predefined jittering (J), we sample from $\mathcal{N}(0,0.01^2)$ with output clipped at $0.05$. We further evaluate state-of-the-art point cloud data augmentation approaches, including PointMixup~\cite{chen2020pointmixup} (PointMixup-Mixup) and PointAugment~\cite{li2020pointaugment} (Sampling-S,T-PointAug.-R,J). We notice from the released code that sampling from predefined scaling and translation is adopted by PointAugment. Finally, we evaluate two variants of our proposed approach: one that includes scaling, translation and rotation operations (AdaPC-S,T,R) and another that further includes jittering (AdaPC-S,T,R,J).

We make the following observations from Tab.\ref{tab:Classification}. First, with predefined augmentation strategies, classification performance is consistently improved compared to the baseline. The improvement is more significant on the dataset scanned from the real world (ScanObjectNN). This is partially explained by the increased diversity of data samples in the real-world dataset. Second, when jittering is naively incorporated, performance drops on ScanObjectNN as seen by comparing Sampling-S,T ($80.20\%$) v.s. Sampling-S,T,J ($76.93\%$). This is consistent with many observations on data augmentation for 3D point clouds that jittering may not be helpful. PointAugment managed to learn jittering policies with heuristic objectives but still failed to improve over the baseline (Sampling-S,T) on ScanObjectNN. In contrast, with a more principled augmentation objective, our AdaPC achieves the best performance on both datasets with S,T,R. We would like to highlight that AdaPC's $0.7\%$ improvement over PointAugment is non-trivial given the limited diversity and highly controlled setting of the ModelNet40 dataset. When jittering is incorporated, AdaPC further improves on ScanObjectNN. Finally, we evaluate AdaPC on two additional backbones, PointNet++ and DGCNN and present results in Tab.~\ref{tab: Classification2}. AdaPC achieves results superior to, if not comparable with, the predefined augmentation baseline, PointMixup and PointAugment.

\begin{table} [!htbp]
  \vspace{-0.13cm}
  \begin{center}
  \setlength\tabcolsep{3pt}
        \resizebox{0.68\linewidth}{!}{
    \begin{tabular}{ccccccc}
    \toprule
          & \multicolumn{1}{r}{} & \multicolumn{1}{r}{} & \multicolumn{1}{r}{} & \multicolumn{1}{r}{} & \multicolumn{1}{c}{ModelNet40} & \multicolumn{1}{c}{ScanObjectNN} \\
    \midrule
          & \multicolumn{2}{c}{Predefined Aug.} & \multicolumn{2}{c}{Learnable Aug.} & \multicolumn{1}{c}{\multirow{2}{*}{\shortstack{Test\\ Acc.}}} & \multicolumn{1}{c}{\multirow{2}{*}{\shortstack{Test\\ Acc.}}} \\
\cmidrule(lr){2-3} \cmidrule(lr){4-5}          & Method & Ops   & Method & Ops   &       &  \\
    \midrule
    \multirow{6}[2]{*}{\begin{sideways}Existing\end{sideways}} & -     & -     & -     & -     & 89.58 & 77.96 \\
          & Sampling  & S, T  & -     & -     & 90.55 & 80.20 \\
          & Sampling  & S, T, R & -     & -     & 90.51 & 77.96 \\
          & Sampling  & S, T, J & -     & -     & 90.88 & 76.93 \\
          & PointMixup & Mixup & -     & -     & 89.90 & - \\
          & Sampling  & S, T  & PointAug. & R, J  & 90.90 & 79.34 \\
    \midrule
    \multirow{2}[2]{*}{\begin{sideways}Ours\end{sideways}} & -     & -     & AdaPC & S, T, R & \textbf{91.61} & 79.86 \\
          & -     & -     & AdaPC & S, T, R, J & 90.80 & \textbf{81.75} \\
    \bottomrule
    \end{tabular}%
    }
    \end{center}
      \caption{Results on 3D point cloud classification. Method and Ops indicate respectively the method adopted for augmentation and the specific operations used for augmentation.}

  \label{tab:Classification}%
\end{table}%

\begin{table} 
  \begin{center}
\scriptsize
  \setlength\tabcolsep{3pt}
        \resizebox{0.62\linewidth}{!}{
\begin{tabular}{cccccc}
    \toprule
          \multicolumn{1}{r}{} & \multicolumn{1}{r}{} & \multicolumn{1}{r}{} & \multicolumn{1}{r}{} & \multicolumn{1}{c}{PointNet++} & \multicolumn{1}{c}{DGCNN} \\
    \midrule
           \multicolumn{2}{c}{Predefined Aug.} & \multicolumn{2}{c}{Learnable Aug.} & \multicolumn{1}{c}{\multirow{2}{*}{\shortstack{Test\\ Acc.}}} & \multicolumn{1}{c}{\multirow{2}{*}{\shortstack{Test\\ Acc.}}} \\
\cmidrule(lr){1-2} \cmidrule(lr){3-4}           Method & Ops   & Method & Ops   &       &  \\
    \midrule
    - &- &- &- &91.41 &91.85 \\
Sampling &S, T, R &- &- &92.26 &92.38 \\
PointMixup &Mixup &- &- &92.70 &- \\
Sampling &S, T &PointAugment* &R, J &92.82 &92.70 \\
\midrule
    - &- &AdaPC &S, T, R &\textbf{92.94} &\textbf{92.82} \\
\bottomrule
\end{tabular}
}
\end{center}
\caption{Additional backbones on ModelNet40 (*for fair comparison, we report the reproduced results for PointAugment under the setting where batch size is 12 and Adam optimizer is used.)}\label{tab: Classification2}
\vspace{-0.6cm}
\end{table}










\subsection{Classification under Non-Fixed Poses}
In the previous setting, the poses of CAD models are all perfectly aligned, thus reducing the effect of data augmentation. Thus, here we consider a more challenging scenario where both training, validation and test data are not perfectly aligned and then investigate the effectiveness of learning the augmentation. Since variations in scaling and translation can often be eliminated by normalization, we consider rotation about the y-axis (gravity) only in this experiment. 

Concretely, we randomly rotate each example by sampling an angle $\theta_r\sim\mathcal{N}(1.0 ,0.2^2)$. We compare AdaPC to vanilla PointNet without augmentation, PointNet using rotation augmentation with angle $\theta\sim\mathcal{U}(-\delta,\delta)$ for several values of $\delta$, PointAugment~\cite{li2020pointaugment} (PointAug.-R,J), a state-of-the-art SO(3)  rotation invariant network VectorNeurons~\cite{deng2021vector} and our final model (AdaPC-S,T,R). The results are shown in Tab.~\ref{tab:NonFixPose}. We observe that including augmentation with a small rotation ($\delta=0.1$) increases baseline model test accuracy from $87.07\%$ (without augmentation) to $89.54\%$, while large rotations ($\delta=3.14$) harm performance; our results suggest that including small pose perturbations is helpful for generalization. PointAugment produces results close to the best manually defined augmentation but is still outperformed by AdaPC. Finally, we observe a much weaker performance by VectorNeurons despite its robustness to SO(3) rotation.

\begin{table}[!htbp]
\begin{center}
    \begin{minipage}{0.42\textwidth}
  \begin{center}
    \setlength\tabcolsep{3pt}
        \resizebox{1\linewidth}{!}{
    \begin{tabular}{llc}
    \toprule
    \multicolumn{2}{c}{Augmentation} & \multirow{2}{*}{\shortstack{Test\\Acc.}} \\
    \cmidrule{1-2}
    Method & Ops  &  \\
    \midrule
    -     & -     & 87.07 \\
    Sampling  & R ($\theta_r\sim\mathcal{U}(-0.05,0.05))$ & 89.42 \\
    Sampling  & R ($\theta_r\sim\mathcal{U}(-0.1,0.1))$ & 89.54 \\
    Sampling  & R ($\theta_r\sim\mathcal{U}(-0.17,0.17))$ & 88.81 \\
    Sampling  & R ($\theta_r\sim\mathcal{U}(-3.14,3.14))$ & 77.87 \\
    PointAug. & R, J & 89.54 \\
    VectorNeurons & -     & 81.68 \\
    AdaPC & S, T, R & \textbf{90.92} \\
    \bottomrule
    \end{tabular}%
    }
    \end{center}
    \caption{Classification results on ModelNet40 under non-fixed poses. }
  \label{tab:NonFixPose}%
  \end{minipage}\quad
   \begin{minipage}{0.40\textwidth}
\begin{center}
    \includegraphics[width=1\linewidth]{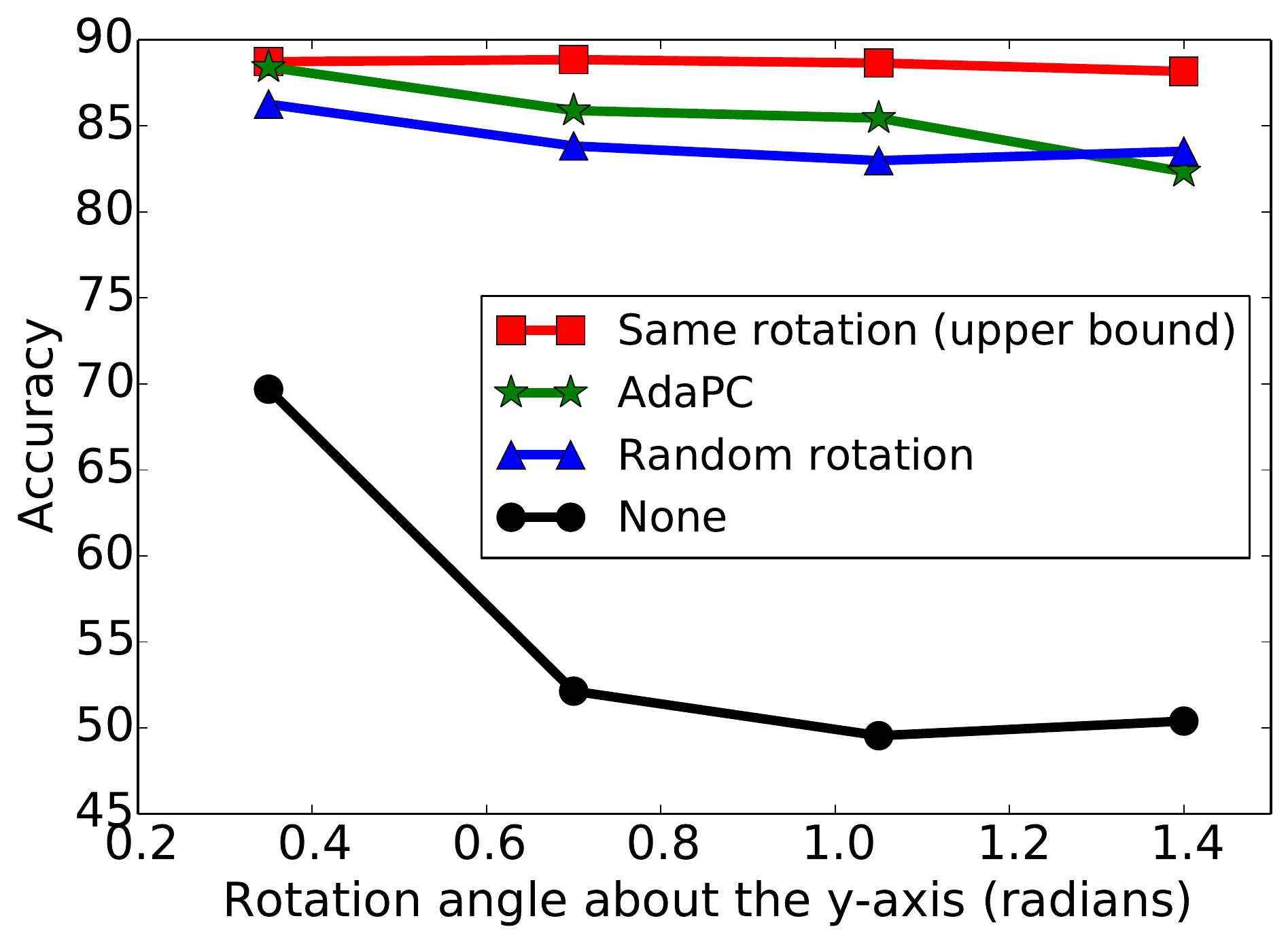}
\end{center}
\vspace{-0.8cm}
\captionof{figure}{Test accuracy on ModelNet40 at various test rotations}
\label{fig:rotation_plot}
  \end{minipage}
 \end{center}
\end{table}%
  \vspace{-0.6cm}

\subsection{Pose Mismatch between Training and Testing}
In this section, we simulate the challenging setting where there is a pose mismatch between training and validation/testing sets, which may occur in practice as detecting and aligning poses for 3D point clouds is a non-trivial task \cite{Wang_2019_CVPR}. We rotate the test and validation sets by 0.35/0.70/1.05/1.4 radians about the y-axis while the training set remains unchanged. We note that the validation and test sets should be drawn from the same distribution for bilevel optimization to work. Here we assume no prior knowledge on rotation angles and hence remove the regularization term. We evaluate 4 methods in this experiment: 1) the ``None'' baseline that does not apply any augmentation, 2) ``Same rotation'', that applies the ground-truth rotation to the training data and thus serves as an upper bound, 3) ``Random rotation'', that applies a random rotation $\theta_r\sim\mathcal{U}(-3.14,3.14)$ to the training data, and 4) AdaPC with only rotation augmentation.
Fig.~\ref{fig:rotation_plot} shows the test accuracy on ModelNet40 at various rotation angles. We see that AdaPC (green line) largely outperforms ``Random rotation'' and is always much better than the no-augmentation ``None'' baseline.



\begin{figure}[!htbp]
\begin{center}
    \includegraphics[width=0.9\linewidth]{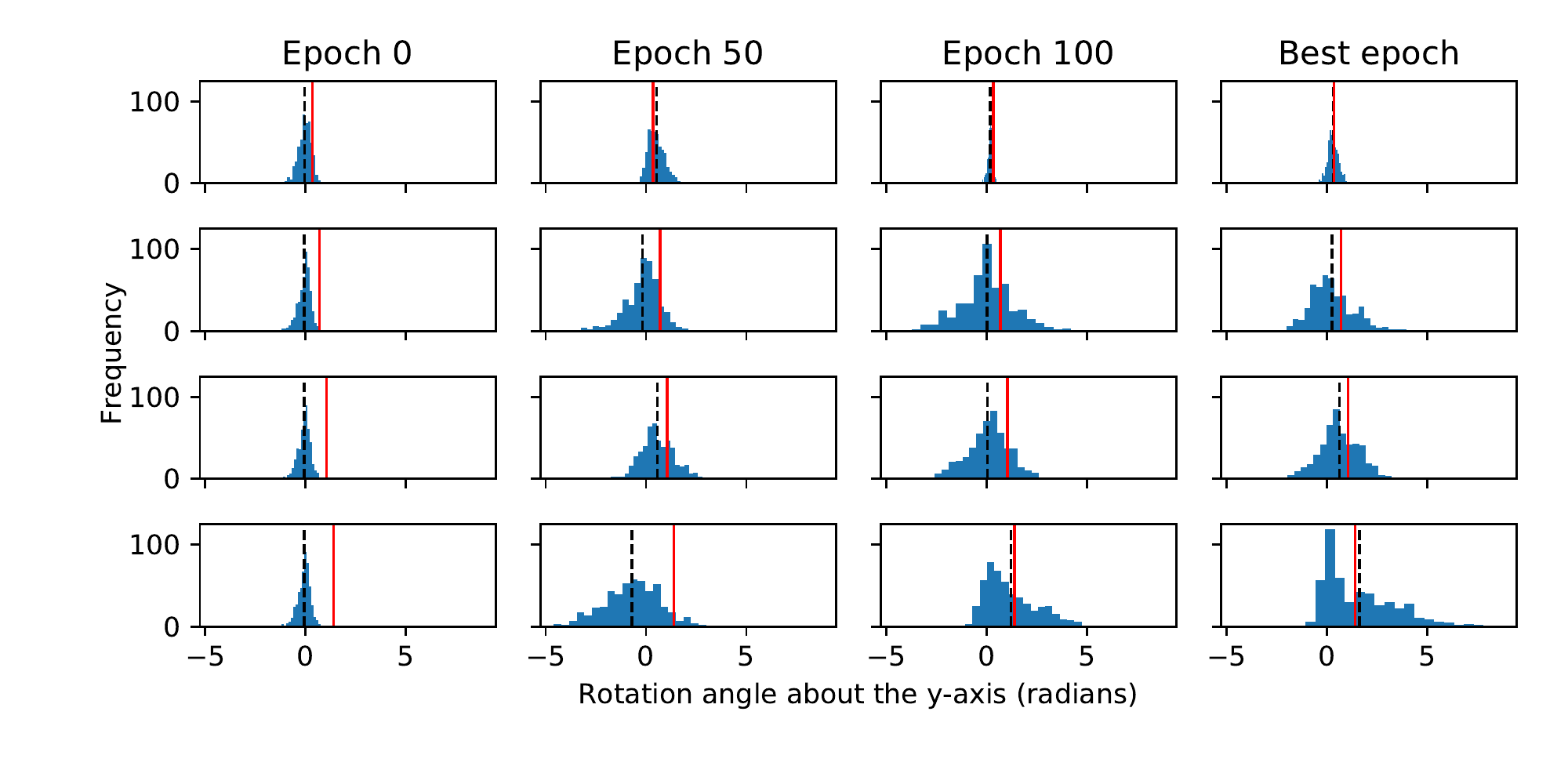}
\caption{Rotation distributions learned by the augmentor over training epochs. The dashed black vertical line indicates the mean of the learned distribution and the red vertical line indicates the ground-truth rotation applied to the test set. From top to bottom, a 0.35/0.70/1.05/1.4 radian rotation is respectively applied to the validation/test sets.}
\label{fig:rotation_distribution}
\end{center}
\end{figure}

We further provide insight into the distribution learned by the proposed augmentor by visualizing the augmentor's output distribution over the course of training in Fig.~\ref{fig:rotation_distribution}. At the beginning, the augmentor's output closely resembles a Gaussian distribution. However, as training progresses, the mean of the augmentor distribution converges to the ground-truth pose. The spread of the distribution at the best epoch suggests that randomness is essential to the success of augmentation.

We also explored if the augmentor can be trained using a subset of the training data and transferred to the full dataset under this setting, as this could accelerate training of the augmentor. Results from this experiment are discussed in Section S2 of the supplementary material and suggest the potential viability of such an approach.

\begin{table}[!htbp]
\vspace{1.0cm}
  \begin{center}
   \setlength\tabcolsep{3pt}
        \resizebox{0.62\linewidth}{!}{
    \begin{tabular}{cccccccccc}
    \toprule
    \multicolumn{1}{c|}{\multirow{4}[2]{*}{\begin{sideways}Aug. Ops\end{sideways}}} & S     & -     & \checkmark     & -     & -    & -  & \checkmark     & \checkmark     & \checkmark \\
    \multicolumn{1}{c|}{} & T     & -     & -     & \checkmark  & -   & -     & \checkmark     & \checkmark     & \checkmark \\
    \multicolumn{1}{c|}{} & R     & -     & -     & -     & \checkmark   & -  & -     & \checkmark     & \checkmark \\
    \multicolumn{1}{c|}{} & J     & -     & -     & -     & -     & \checkmark  & -    & -     & \checkmark \\
    \midrule
    \multicolumn{2}{c}{ModelNet40} & 89.58 & 90.19 & 90.15 & 90.15 &89.82 &90.84 & \textbf{91.61} & 90.80 \\
    \multicolumn{2}{c}{ScanObjectNN} & 77.96 & 77.62 & 80.72 & 75.59 &76.62 &78.14 & 79.86 & \textbf{81.75} \\
    \bottomrule
    \end{tabular}%
    }
    \end{center}
    
  \label{tab:ablation}%

  \caption{Ablation study on individual augmentation operations.}

\end{table}%

\subsection{Ablation Study}
\subsubsection{Different combinations of augmentation operations}
Here we explore the effectiveness of various combinations of augmentation operations on both ModelNet40 and ScanObjectNN datasets. From the results in Tab.~\ref{tab:ablation}, we see that using each of scaling, translation, rotation, and jittering alone outperforms the no-augmentation baseline on ModelNet40, and that the best results are achieved by combining the first 3 types of augmentations ($91.61\%$ on ModelNet40). We observe a slight drop in performance on ModelNet40 when further including jittering. One possible reason for this drop in performance is that ModelNet40 consists of points uniformly sampled from CAD models that are noise-free. Hence, training samples augmented with jittering deviate from the distribution of noise-free test samples, causing a drop in performance. In contrast, ScanObjectNN is cropped from real-world scenes, thus shapes are naturally contaminated by noise and jittering is more effective. Our observations suggest that research into domain-specific augmentation strategies will be valuable.
\vspace{-0.3cm}

\vspace{-0.3cm}
\section{Conclusion}
Data augmentation plays an important role for deep learning on 3D point clouds. Different from existing heuristic augmentation methods, we propose an automatic data augmentation method, AdaPC,  based on a more principled bilevel optimization framework. In AdaPC, an augmentor network is used to predict transformation parameters for augmentation. This augmentor is then learned by minimizing the validation loss. AdaPC achieved very competitive results on two point cloud classification datasets. We also demonstrate that AdaPC is effective in a pose mismatch scenario, where the mean of the learned augmentation distribution converged to the true pose. Our work suggests that future works on point cloud augmentation should include evaluations on challenging real-world datasets and settings to demonstrate the true value of data augmentation. \\

\vspace{0.5cm}
\noindent\textbf{Acknowledgement}: This research is supported by the Agency for Science, Technology and Research, Singapore (A*STAR) under its Career Development Award (Grant No. 202D8243). Discussions with Xiatian Zhu are gratefully acknowledged.

\break
\bibliography{egbib}

\begin{thebibliography}{33}
\providecommand{\natexlab}[1]{#1}
\providecommand{\url}[1]{\texttt{#1}}
\expandafter\ifx\csname urlstyle\endcsname\relax
  \providecommand{\doi}[1]{doi: #1}\else
  \providecommand{\doi}{doi: \begingroup \urlstyle{rm}\Url}\fi

\bibitem[Antoniou et~al.(2017)Antoniou, Storkey, and Edwards]{antoniou2017data}
Antreas Antoniou, Amos Storkey, and Harrison Edwards.
\newblock Data augmentation generative adversarial networks.
\newblock \emph{arXiv preprint arXiv:1711.04340}, 2017.

\bibitem[Atzmon et~al.(2018)Atzmon, Maron, and Lipman]{atzmon2018point}
Matan Atzmon, Haggai Maron, and Yaron Lipman.
\newblock Point convolutional neural networks by extension operators.
\newblock \emph{ACM Transactions on Graphics}, 2018.

\bibitem[Chen et~al.(2020)Chen, Hu, Gavves, Mensink, Mettes, Yang, and
  Snoek]{chen2020pointmixup}
Yunlu Chen, Vincent~Tao Hu, Efstratios Gavves, Thomas Mensink, Pascal Mettes,
  Pengwan Yang, and Cees~GM Snoek.
\newblock Pointmixup: Augmentation for point clouds.
\newblock In \emph{European Conference on Computer Vision}, 2020.

\bibitem[Cheng et~al.(2020)Cheng, Leng, Cubuk, Zoph, Bai, Ngiam, Song, Caine,
  Vasudevan, Li, et~al.]{cheng2020improving}
Shuyang Cheng, Zhaoqi Leng, Ekin~Dogus Cubuk, Barret Zoph, Chunyan Bai, Jiquan
  Ngiam, Yang Song, Benjamin Caine, Vijay Vasudevan, Congcong Li, et~al.
\newblock Improving 3d object detection through progressive population based
  augmentation.
\newblock In \emph{European Conference on Computer Vision}, 2020.

\bibitem[Choi et~al.(2020)Choi, Song, and Kwak]{choi2020part}
Jaeseok Choi, Yeji Song, and Nojun Kwak.
\newblock Part-aware data augmentation for 3d object detection in point cloud.
\newblock \emph{arXiv preprint arXiv:2007.13373}, 2020.

\bibitem[Cubuk et~al.(2018)Cubuk, Zoph, Mane, Vasudevan, and
  Le]{cubuk2018autoaugment}
Ekin~D Cubuk, Barret Zoph, Dandelion Mane, Vijay Vasudevan, and Quoc~V Le.
\newblock Autoaugment: Learning augmentation policies from data.
\newblock In \emph{IEEE/CVF Conference on Computer Vsion and Pattern
  Recognition}, 2018.

\bibitem[Cubuk et~al.(2020)Cubuk, Zoph, Shlens, and Le]{cubuk2020randaugment}
Ekin~D Cubuk, Barret Zoph, Jonathon Shlens, and Quoc~V Le.
\newblock Randaugment: Practical automated data augmentation with a reduced
  search space.
\newblock In \emph{Proceedings of the IEEE/CVF Conference on Computer Vision
  and Pattern Recognition Workshops}, 2020.

\bibitem[Deng et~al.(2021)Deng, Litany, Duan, Poulenard, Tagliasacchi, and
  Guibas]{deng2021vector}
Congyue Deng, Or~Litany, Yueqi Duan, Adrien Poulenard, Andrea Tagliasacchi, and
  Leonidas Guibas.
\newblock Vector neurons: A general framework for so (3)-equivariant networks.
\newblock \emph{arXiv preprint arXiv:2104.12229}, 2021.

\bibitem[Goodfellow et~al.(2014)Goodfellow, Pouget-Abadie, Mirza, Xu,
  Warde-Farley, Ozair, Courville, and Bengio]{goodfellow2014generative}
Ian~J Goodfellow, Jean Pouget-Abadie, Mehdi Mirza, Bing Xu, David Warde-Farley,
  Sherjil Ozair, Aaron~C Courville, and Yoshua Bengio.
\newblock Generative adversarial nets.
\newblock In \emph{NIPS}, 2014.

\bibitem[Lee et~al.(2021)Lee, Lee, Lee, Lee, Lee, Woo, and
  Lee]{lee2021regularization}
Dogyoon Lee, Jaeha Lee, Junhyeop Lee, Hyeongmin Lee, Minhyeok Lee, Sungmin Woo,
  and Sangyoun Lee.
\newblock Regularization strategy for point cloud via rigidly mixed sample.
\newblock In \emph{Proceedings of the IEEE/CVF Conference on Computer Vision
  and Pattern Recognition}, 2021.

\bibitem[Lemley et~al.(2017)Lemley, Bazrafkan, and Corcoran]{lemley2017smart}
Joseph Lemley, Shabab Bazrafkan, and Peter Corcoran.
\newblock Smart augmentation learning an optimal data augmentation strategy.
\newblock \emph{IEEE Access}, 2017.

\bibitem[Li et~al.(2020{\natexlab{a}})Li, Li, Heng, and Fu]{li2020pointaugment}
Ruihui Li, Xianzhi Li, Pheng-Ann Heng, and Chi-Wing Fu.
\newblock Pointaugment: an auto-augmentation framework for point cloud
  classification.
\newblock In \emph{Proceedings of the IEEE/CVF Conference on Computer Vision
  and Pattern Recognition}, 2020{\natexlab{a}}.

\bibitem[Li et~al.(2020{\natexlab{b}})Li, Hu, Wang, Hospedales, Robertson, and
  Yang]{li2020differentiable}
Yonggang Li, Guosheng Hu, Yongtao Wang, Timothy Hospedales, Neil~M Robertson,
  and Yongxin Yang.
\newblock Differentiable automatic data augmentation.
\newblock In \emph{European Conference on Computer Vision}, 2020{\natexlab{b}}.

\bibitem[Lim et~al.(2019)Lim, Kim, Kim, Kim, and Kim]{NEURIPS2019_6add07cf}
Sungbin Lim, Ildoo Kim, Taesup Kim, Chiheon Kim, and Sungwoong Kim.
\newblock Fast autoaugment.
\newblock In \emph{Advances in Neural Information Processing Systems}, 2019.

\bibitem[Liu et~al.(2018)Liu, Simonyan, and Yang]{liu2018darts}
Hanxiao Liu, Karen Simonyan, and Yiming Yang.
\newblock Darts: Differentiable architecture search.
\newblock In \emph{International Conference on Learning Representations}, 2018.

\bibitem[Mai et~al.(2021)Mai, Hu, Chen, Shen, and Shen]{mai2021metamixup}
Zhijun Mai, Guosheng Hu, Dexiong Chen, Fumin Shen, and Heng~Tao Shen.
\newblock Metamixup: Learning adaptive interpolation policy of mixup with
  metalearning.
\newblock \emph{IEEE Transactions on Neural Networks and Learning Systems},
  2021.

\bibitem[Mirza and Osindero(2014)]{mirza2014conditional}
Mehdi Mirza and Simon Osindero.
\newblock Conditional generative adversarial nets.
\newblock \emph{arXiv preprint arXiv:1411.1784}, 2014.

\bibitem[Mounsaveng et~al.(2021)Mounsaveng, Laradji, Ben~Ayed, Vazquez, and
  Pedersoli]{mounsaveng2021learning}
Saypraseuth Mounsaveng, Issam Laradji, Ismail Ben~Ayed, David Vazquez, and
  Marco Pedersoli.
\newblock Learning data augmentation with online bilevel optimization for image
  classification.
\newblock In \emph{Proceedings of the IEEE/CVF Winter Conference on
  Applications of Computer Vision}, 2021.

\bibitem[Mun et~al.(2017)Mun, Park, Han, and Ko]{mun2017generative}
Seongkyu Mun, Sangwook Park, David~K Han, and Hanseok Ko.
\newblock Generative adversarial network based acoustic scene training set
  augmentation and selection using svm hyper-plane.
\newblock \emph{Proc. DCASE}, 2017.

\bibitem[Perez and Wang(2017)]{perez2017effectiveness}
Luis Perez and Jason Wang.
\newblock The effectiveness of data augmentation in image classification using
  deep learning.
\newblock \emph{Convolutional Neural Networks Vis. Recognit}, 2017.

\bibitem[Phan et~al.(2018)Phan, Le~Nguyen, Nguyen, and Bui]{phan2018dgcnn}
Anh~Viet Phan, Minh Le~Nguyen, Yen Lam~Hoang Nguyen, and Lam~Thu Bui.
\newblock Dgcnn: A convolutional neural network over large-scale labeled
  graphs.
\newblock \emph{Neural Networks}, 2018.

\bibitem[Qi et~al.(2017{\natexlab{a}})Qi, Su, Mo, and Guibas]{qi2017pointnet}
Charles~R Qi, Hao Su, Kaichun Mo, and Leonidas~J Guibas.
\newblock Pointnet: Deep learning on point sets for 3d classification and
  segmentation.
\newblock In \emph{Proceedings of the IEEE conference on computer vision and
  pattern recognition}, 2017{\natexlab{a}}.

\bibitem[Qi et~al.(2017{\natexlab{b}})Qi, Yi, Su, and Guibas]{qi2017pointnet++}
Charles~R Qi, Li~Yi, Hao Su, and Leonidas~J Guibas.
\newblock Pointnet++: Deep hierarchical feature learning on point sets in a
  metric space.
\newblock In \emph{Advances in Neural Information Processing Systems},
  2017{\natexlab{b}}.

\bibitem[Tanaka and Aranha(2019)]{tanaka2019data}
Fabio Henrique Kiyoiti dos~Santos Tanaka and Claus Aranha.
\newblock Data augmentation using gans.
\newblock \emph{arXiv preprint arXiv:1904.09135}, 2019.

\bibitem[Thomas et~al.(2019)Thomas, Qi, Deschaud, Marcotegui, Goulette, and
  Guibas]{thomas2019kpconv}
Hugues Thomas, Charles~R Qi, Jean-Emmanuel Deschaud, Beatriz Marcotegui,
  Fran{\c{c}}ois Goulette, and Leonidas~J Guibas.
\newblock Kpconv: Flexible and deformable convolution for point clouds.
\newblock In \emph{Proceedings of the IEEE/CVF International Conference on
  Computer Vision}, 2019.

\bibitem[Uy et~al.(2019)Uy, Pham, Hua, Nguyen, and Yeung]{uy2019revisiting}
Mikaela~Angelina Uy, Quang-Hieu Pham, Binh-Son Hua, Thanh Nguyen, and Sai-Kit
  Yeung.
\newblock Revisiting point cloud classification: A new benchmark dataset and
  classification model on real-world data.
\newblock In \emph{Proceedings of the IEEE/CVF International Conference on
  Computer Vision}, 2019.

\bibitem[Verma et~al.(2019)Verma, Lamb, Beckham, Najafi, Mitliagkas, Lopez-Paz,
  and Bengio]{verma2019manifold}
Vikas Verma, Alex Lamb, Christopher Beckham, Amir Najafi, Ioannis Mitliagkas,
  David Lopez-Paz, and Yoshua Bengio.
\newblock Manifold mixup: Better representations by interpolating hidden
  states.
\newblock In \emph{International Conference on Machine Learning}, 2019.

\bibitem[Wang et~al.(2019)Wang, Sridhar, Huang, Valentin, Song, and
  Guibas]{Wang_2019_CVPR}
He~Wang, Srinath Sridhar, Jingwei Huang, Julien Valentin, Shuran Song, and
  Leonidas~J. Guibas.
\newblock Normalized object coordinate space for category-level 6d object pose
  and size estimation.
\newblock In \emph{Proceedings of the IEEE/CVF Conference on Computer Vision
  and Pattern Recognition}, 2019.

\bibitem[Wu et~al.(2015)Wu, Song, Khosla, Yu, Zhang, Tang, and Xiao]{wu20153d}
Zhirong Wu, Shuran Song, Aditya Khosla, Fisher Yu, Linguang Zhang, Xiaoou Tang,
  and Jianxiong Xiao.
\newblock 3d shapenets: A deep representation for volumetric shapes.
\newblock In \emph{Proceedings of the IEEE conference on computer vision and
  pattern recognition}, 2015.

\bibitem[Yoo et~al.(2020)Yoo, Ahn, and Sohn]{yoo2020rethinking}
Jaejun Yoo, Namhyuk Ahn, and Kyung-Ah Sohn.
\newblock Rethinking data augmentation for image super-resolution: A
  comprehensive analysis and a new strategy.
\newblock In \emph{Proceedings of the IEEE/CVF Conference on Computer Vision
  and Pattern Recognition}, 2020.

\bibitem[Yun et~al.(2019)Yun, Han, Oh, Chun, Choe, and Yoo]{yun2019cutmix}
Sangdoo Yun, Dongyoon Han, Seong~Joon Oh, Sanghyuk Chun, Junsuk Choe, and
  Youngjoon Yoo.
\newblock Cutmix: Regularization strategy to train strong classifiers with
  localizable features.
\newblock In \emph{Proceedings of the IEEE/CVF International Conference on
  Computer Vision}, 2019.

\bibitem[Zhang et~al.(2017)Zhang, Cisse, Dauphin, and
  Lopez-Paz]{zhang2017mixup}
Hongyi Zhang, Moustapha Cisse, Yann~N Dauphin, and David Lopez-Paz.
\newblock mixup: Beyond empirical risk minimization.
\newblock In \emph{International Conference on Learning Representations}, 2017.

\bibitem[Zhang et~al.(2021)Zhang, Chen, Ouyang, Liu, Zhu, Chen, Meng, and
  Wu]{zhang2021pointcutmix}
Jinlai Zhang, Lvjie Chen, Bo~Ouyang, Binbin Liu, Jihong Zhu, Yujing Chen,
  Yanmei Meng, and Danfeng Wu.
\newblock Pointcutmix: Regularization strategy for point cloud classification.
\newblock \emph{arXiv preprint arXiv:2101.01461}, 2021.

\end{thebibliography}



\title{On Automatic Data Augmentation for 3D Point Cloud Classification (Supplementary Material)}

\renewcommand*{\thesection}{S\arabic{section}}
\renewcommand*{\thesubsection}{S\arabic{subsection}}
\renewcommand*{\thesubsubsection}{S\arabic{subsection}.\arabic{subsubsection}}
\renewcommand*{\thefigure}{S\arabic{figure}}
\renewcommand*{\thetable}{S\arabic{table}}
\section*{\LARGE On Automatic Data Augmentation for 3D Point Cloud Classification (Supplementary Material)}

In this supplementary material, we further provide insight into augmentor designs and whether we can more efficiently learn the augmentor using only a subset of the data.

\setcounter{section}{0}
\setcounter{figure}{0}
\setcounter{table}{0}
\section{Alternative augmentor designs}\label{sect:alternative_augmentor}
In this section, we explore the effect of alternative parameterizations of the augmentor. Specifically, we consider parametric probability distributions -- uniform and Gaussian distributions -- in our experiments. We first evaluate a uniform distribution augmentor where the lower bound $lb$ and upper bound $ub$ are the learnable parameters. Regularization is applied to encourage their mean to be close to a predefined constant $\hat{\theta}$: $||(lb+ub)/2-\hat{\theta}||_2$. In addition, we evaluate a Gaussian distribution augmentor where  $\mu$ and standard deviation $\sigma$ are the learnable parameters. Regularization is applied only to the mean: $||\mu-\hat{\theta}||_2$. For fair comparison, we apply regularization with a weight of 0.5 for all augmentors. 
The results on ModelNet40 classification are presented in Tab.~\ref{tab:augmentor}. We observe that AdaPC with Gaussian distribution and neural network augmentor outperforms the no-augmentor baseline. More importantly, only AdaPC with the neural network augmentor achieves better performance than predefined augmentations (second row in Tab.~\ref{tab:augmentor}), suggesting that the more flexible neural network parametrization for the augmentor is essential to the success of learning stronger augmentations.


\begin{table}[!htp]\centering
\scriptsize
\begin{tabular}{lrrrrr}\toprule
Augmentor &Augmentor params &Fixed ops &Learnable ops &Test accuracy \\\midrule
None &None &None &None &89.58 \\
None &None &S, T, Ry &None &90.51 \\
Uniform &lower and upper bound &None &S, T, Ry &89.46 \\
Gaussian &mean and standard deviation &None &S, T, Ry &90.39 \\
AdaPC &neural networks &None &S, T, Ry &91.61 \\
\bottomrule
\end{tabular}
\caption{Alternative augmentor results on ModelNet40}\label{tab:augmentor}
\end{table}

\begin{table}[!h]\centering
\scriptsize
\begin{tabular}{lrrrrr}\toprule
Augmentor &Fixed ops &Learnable ops &Dataset for meta training &Test accuracy \\\midrule
None &None &None &None &82.14 \\
None &Ry = 0.7 &None &None &88.85 \\
AdaPC &None &Ry &ModelNet40 &85.89 \\
AdaPC &None &Ry &10\% ModelNet40 &88.33 \\
\bottomrule
\end{tabular}
\caption{Training the augmentor on a subset of ModelNet40.}\label{tab:transfer}
\end{table}

\section{Learning the augmentor on a subset of data}
In this section, we accelerate the learning process by training the augmentor using only a subset of the data. Concretely, we train the augmentor on a small subset ($10\%$ of all available data) of ModelNet40 and then use the learnt augmentor with the best validation accuracy to re-train a classifier with all available data. During re-training, the augmentor is fixed and only the classifier is trainable and we follow the same training protocol. In this experiment,  both validation and test sets are rotated by 0.7 radian about the y-axis. We observe from Tab.~\ref{tab:transfer} that AdaPC achieves a very competitive result (last row, $88.33\%$) when trained on $10\%$ of ModelNet40 and transferred to the full dataset, suggesting that it is viable to use this strategy to accelerate augmentor learning. More specifically, we can train the augmentor 10 times faster with 10\% ModelNet40, and the augmentor adds a negligible computational overhead in the re-training stage on the full dataset. 


\end{document}